\pdfoutput=1

\documentclass[11pt]{article}

\usepackage[]{ACL2023}

\usepackage{times}
\usepackage{latexsym}

\usepackage[T1]{fontenc}

\usepackage[utf8]{inputenc}

\usepackage{microtype}

\usepackage{inconsolata}
\usepackage{enumitem}
\usepackage{multirow}
\usepackage{booktabs}
\usepackage{arydshln}
\usepackage{graphicx}
\usepackage{CJKutf8}
\usepackage{makecell}
\usepackage{amsmath}
\usepackage{amssymb}

\definecolor{forestgreen}{rgb}{0.13, 0.55, 0.13}

\title{Findings of the WMT 2023 Shared Task on Discourse-Level Literary Translation: A Fresh Orb in the Cosmos of LLMs}

\author{Longyue Wang, Zhaopeng Tu, Yan Gu, Siyou Liu, Dian Yu, Qingsong Ma,\\ \bf Chenyang Lyu, Liting Zhou, Chao-Hong Liu, Yufeng Ma, Weiyu Chen, \\ \bf Yvette Graham, Bonnie Webber, Philipp Koehn, Andy Way, Yulin Yuan, Shuming Shi\\ vinnylywang@tencent.com}

\begin{document}
\maketitle
\begin{abstract}
Translating literary works has perennially stood as an elusive dream in machine translation (MT), a journey steeped in intricate challenges. To foster progress in this domain, we hold a new shared task at WMT 2023, the first edition of the {\em Discourse-Level Literary Translation}. First, we (Tencent AI Lab and China Literature Ltd.) release a copyrighted and document-level Chinese-English web novel corpus. Furthermore, we put forth an industry-endorsed criteria to guide human evaluation process. This year, we totally received 14 submissions from 7 academia and industry teams. We employ both automatic and human evaluations to measure the performance of the submitted systems. The official ranking of the systems is based on the overall human judgments. In addition, our extensive analysis reveals a series of interesting findings on literary and discourse-aware MT. We release data, system outputs, and leaderboard at 
\url{http://www2.statmt.org/wmt23/literary-translation-task.html}.
\end{abstract}

\section{Introduction}

In past decades, the evolution of machine translation (MT) has undergone significant improvements in accuracy and efficiency, leading to many practical applications in various fields~\cite{bojar-etal-2014-findingss,barrault-etal-2019-findingss,farhad2021findings,kocmi2022findings}. Despite its success, MT still struggles in certain intricate scenarios to deliver translations that meet high standards~\cite{laubli2018has,koehn2017six}. Translating literary texts is considered to be the greatest challenge for MT due to its complex nature~\cite{toral2018level,toral2018post,ghazvininejad2018neural}:

\begin{itemize}[leftmargin=*,topsep=0.1em,itemsep=0.1em,parsep=0.1em]

\item {\em Rich Linguistic and Cultural Phenomena}: literary texts contain more complex linguistic and cultural knowledge than non-literary ones~\cite{voigt2012towards,ghazvininejad2018neural}. To generate a cohesive and coherent output, MT models require an understanding of the intended meaning and structure of the text at discourse level \cite{wang2016novel,wang2018translating,wang2018learning,wang2019one,wang2023survey}. Furthermore, it demands skillful adaptation of cultural references, idioms, and subtle expressions to capture the essence of the original work in target languages.

\item {\em Limited Data}: existing document-level datasets are news articles and technical documents~\cite{liu2020corpora,thai2022exploring}; there is limited availability of copyrighted, discourse-level, parallel data in the literature domain. This makes it difficult to develop models that are able to handle the complexities of literary translation.

\item {\em Long-Range Context}: literature such as novels have much longer contexts than texts in other domains (e.g. news articles). Translation models need to acquire the capacity of modeling long-range context for learning translation consistency and lexical choice~\cite{wang2017exploiting,wang2019discourse,matusov2019challenges,du2023on}. 

\item {\em Unreliable Evaluation Methods}: literary evaluation needs to measure the meaning and structure of the text, and the nuances and complexities of the source language. A single automatic evaluation using a single reference is unreliable. Thus, professional translators with well-defined error typologies and targeted automatic evaluation are considered a complement~\cite{matusov2019challenges}.

\end{itemize}

With the swift progression of MT and the notable advancements in Large Language Models (LLM)~\cite{ouyang2022training,openai2023gpt4}, our curiosity is piqued regarding the efficacy of MT and LLM in the realm of literary translation. We aim to explore the extent to which these technologies can aid in addressing the intricate challenges of translating literary works. Therefore, we hold the first edition of the {\em Discourse-Level Literary Translation} in WMT 2023. 
Literary texts encompass a wide range of forms, including novels, short stories, poetry, plays, essays, and more. Among these, {\em web novels}, also known as online or internet novels, represent a unique and rapidly growing subset of literature. Their popularity, accessibility, and diverse genres set them apart. As they provide not only an extensive volume of text but also exhibit distinctive linguistic features, cultural phenomena, and simulations of societies, web novels can serve as valuable resources and challenging for MT research. 
\begin{figure}[t] 
    \centering
    \includegraphics[width=0.48\textwidth]{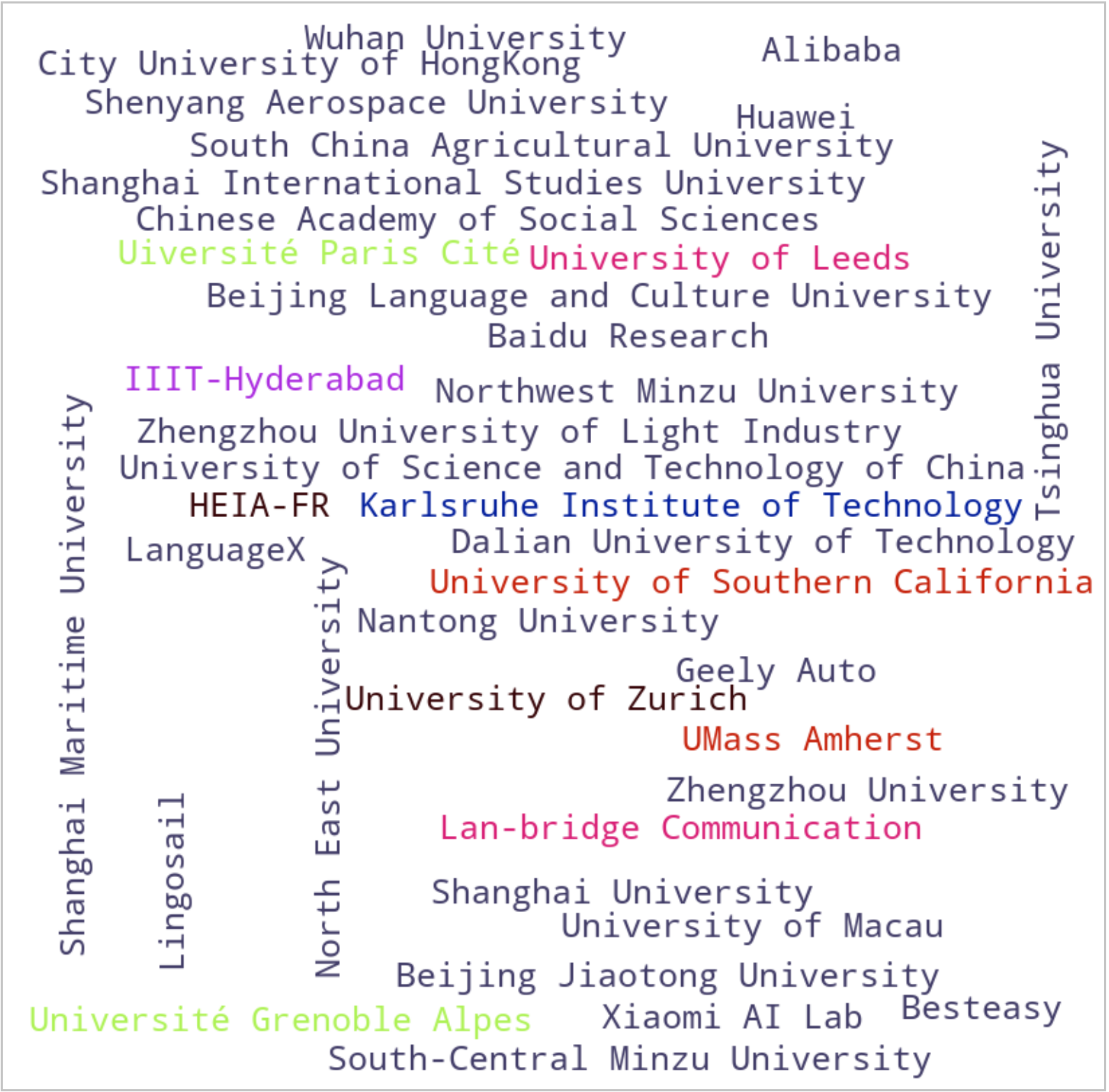}
    \caption{The word cloud represents institute and companies from different regions that downloaded the GuoFeng Webnovel Corpus.} 
    \label{fig:wordcloud}
\end{figure}
This year, the shared task mainly focuses on {\em document-level web novels}, and we introduce a document-level benchmark dataset and establish human evaluation criteria specifically tailored to address the challenges of literary translation:
\begin{itemize}[leftmargin=*,topsep=0.1em,itemsep=0.1em,parsep=0.1em]

\item {\bf Benchmark Dataset}: We build and release a copyrighted and high-quality Chinese-English training corpus, comprising 2 million sentences sourced from 179 web fictions. This dataset preserves both book-level and chapter-level contexts, and features manually-aligned sentence pairs. We also provide three types of testsets, varying in distribution and document length (in Section~\ref{sec:2}).

\item  {\bf Evaluation Methods}: In order to evaluate the translation quality of the participating systems we used both automatic and human evaluation methods. About automatic evaluation, we employ document-level sacreBLEU (d-BLEU) as our metric, which is computed by matching n-grams in the whole document~\cite{liu2020multilingual,post2018call}. In terms of human evaluation, we propose a well-defined criteria by adapting multidimensional quality metrics (MQM) \cite{lommel2014multidimensional} to fit the context of literary translation. Note that all evaluations are case-sensitive (in Section~\ref{sec:3}).

\end{itemize}

\noindent We introduce the task overview and submission form in Section~\ref{sec:4}. This year, 14 submissions were received from 7 different teams, which are detailed in Section~\ref{sec:5}. We report the evaluation results in Section~\ref{sec:6} followed by the conclusion in Section~\ref{sec:7}.

\begin{figure}[t] 
    \centering
    \includegraphics[width=0.48\textwidth]{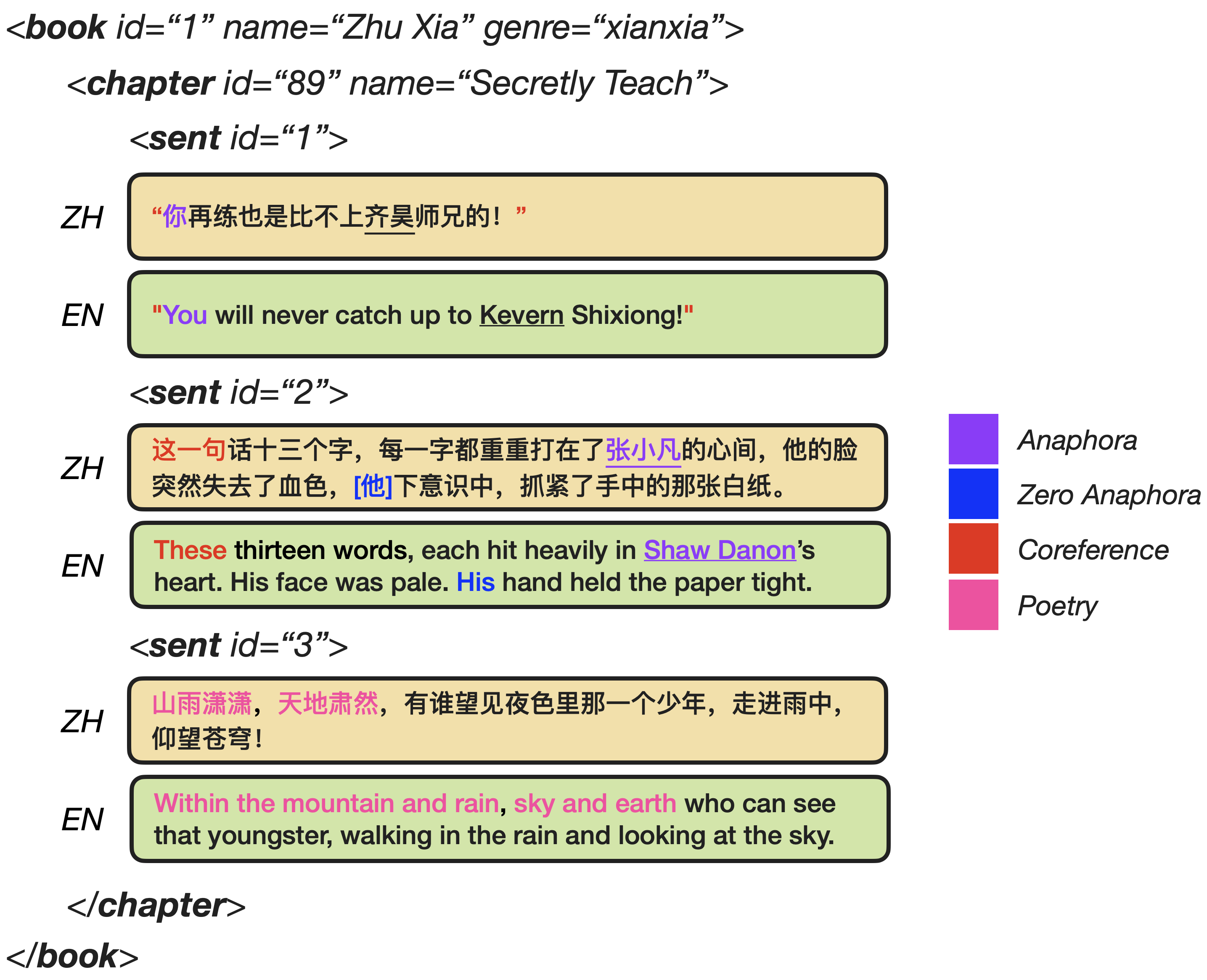}
    \caption{Illustration of discourse-level literary translation, which is sampled from our Web Fiction Corpus. Colored words demonstrate rich linguistic phenomena.} 
    \label{fig:example}
\end{figure}

\section{The GuoFeng Webnovel Corpus}
\label{sec:2}

We release a copyrighted and high-quality Chinese-English corpus on web novels. Additionally, we provide in-domain pretrained models as supplementary resources. As shown in Figure~\ref{fig:wordcloud}, a total of 45 institutes and companies from various regions have downloaded our dataset, showing that the prposed tasks and data have garnered widespread interest.

\subsection{Datasets}
\label{sec:2-1}

\paragraph{Copyright}

Copyright is a crucial consideration when it comes to releasing literary texts, and it is also one of the primary reasons for limiting the scale of data in this domain. 
We, Tencent AI Lab and China Literature Ltd., are the copyright owners of the web fictions included in this dataset. In order to promote the advancement of research in this field, we make this data available to the research community, subject to certain terms and conditions. 
\begin{itemize}[leftmargin=*,topsep=0.1em,itemsep=0.1em,parsep=0.1em]

\item After registration, WMT participants can use the corpus for non-commercial research purposes and follow the principle of fair use (CC-BY).

\item Modifying or redistributing the dataset is strictly prohibited. 

\item You should cite the this paper and claim the original download link.

\end{itemize}

\begin{figure*}

\begin{minipage}[b]{0.5\linewidth}
    \begin{tabular}{l rrrrr}
    \toprule
    \textbf{Dataset} & \textbf{\#Book} & \textbf{\#Chap.} & \textbf{\#Sent.} & \bf \#Word & \textbf{|D|} \\
    \midrule
    Train & 179 & 22.6K & 1.9M & 32.0M & 1.4K\\ \midrule
    Valid$_1$	& 22	&22	&755 & 18.3K & 832\\
    Test$_1$	& 26	&22	&697 & 19.5K & 884\\
    \cdashline{1-6}\noalign{\vskip 0.5ex}
    Valid$_2$	& 10	&10	&853 & 16.0K & 1.6K\\ 
    Test$_2$	& 12	&12	&917 & 16.7K & 1.4K\\ \midrule
    Test$_{final}$	& 12	&239	& 16.7K & 337.0K & $^*$28.1K\\
    \bottomrule
    \end{tabular}
\end{minipage}
\hfill
\begin{minipage}{0.37\textwidth}
  \centering
  \includegraphics[width=1\linewidth]{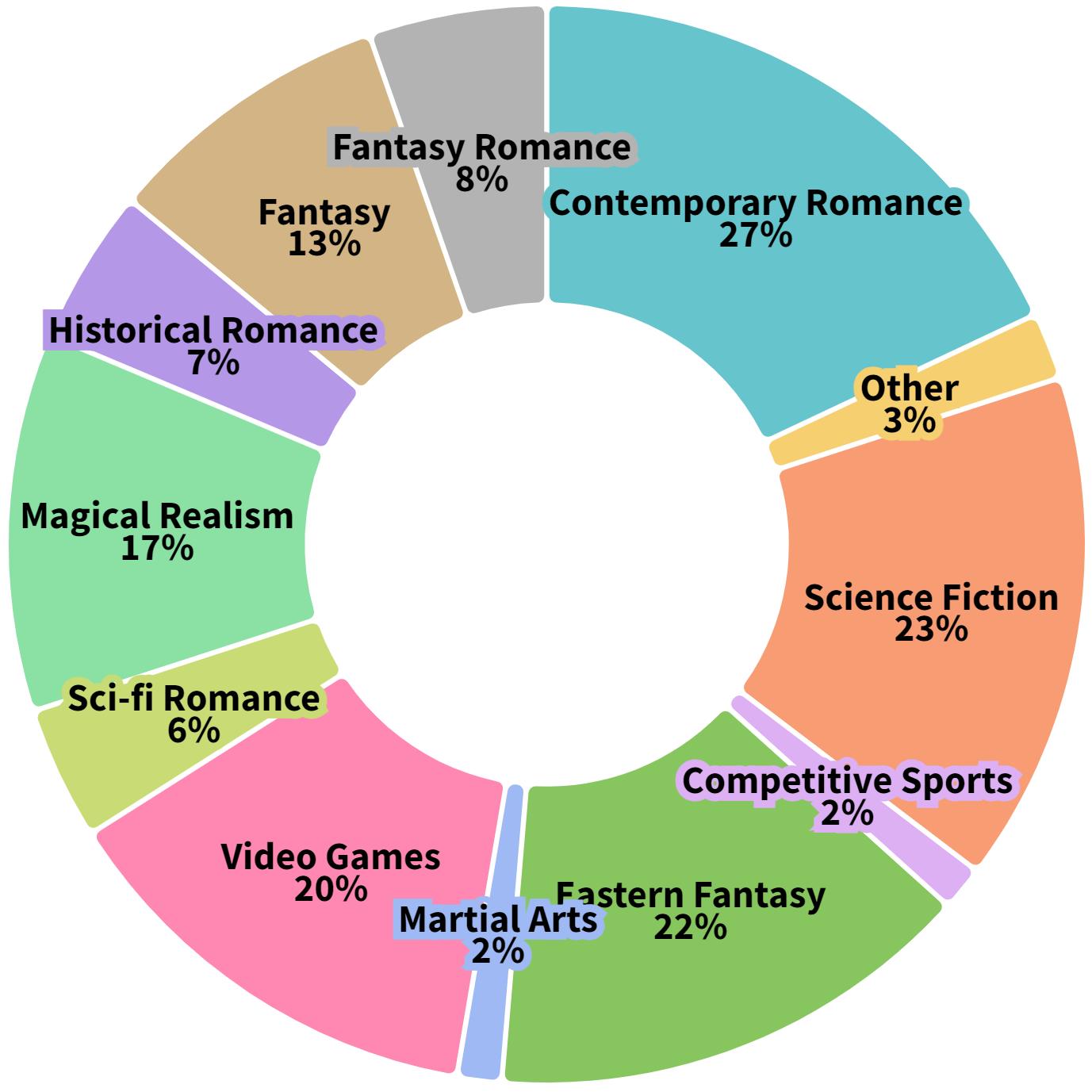}
\end{minipage}
    \caption{\label{tab:data_statics}Data statistics of the GuoFeng Webnovel Corpus on number of book, chapter (\#Chap.), sentence (\#Sent.), word, and genre distribution in training set. The \#Word is based on English texts. For dataset$_1$, books overlap with the training data, whereas dataset$_2$ contains unseen books. Thus, each chapter is treated as a separate document. For Test$_{final}$, around 20 consecutive chapters from each book are selected, treating all chapters within a book as a long document. The document length (|D|) is calculated by dividing \#Word divided by the number of documents. }
\end{figure*}

\paragraph{Data Processing}
The web novels are originally written in Chinese by web novel writers and then translated into English by professional translators. Our data processing involves a combination of automated and manual techniques: 1) we match Chinese books with its English counterparts based on bilingual titles; 2) within each book, Chinese-English chapters are aligned using Chapter ID numbers; 3) within each chapter, we build a MT-based sentence aligner to align sentences in parallel, preserving the sentence order in the chapter; 4) human annotators are engaged to review and correct any discrepancies in sentence-level alignment. To ensure the retention of discourse information, we permit null alignments. We totally spent 6 months addressing copyright issues and around 40,000 euros for human annotation.
Figure \ref{fig:example} shows the final format of our corpus. 

\paragraph{Training/Validation/Testing Data}

Table~\ref{tab:data_statics} lists data statistics of our dataset. As seen, the {\em training set} contains 23K continuous chapters from 179 web novels, covering 14 genres such as fantasy science and romance. 
To enable participants to evaluate model performance by themselves, we provide two {\em unofficial validation/testing sets} with one reference. For {dataset$_1$}, books overlap with the training data, whereas {dataset$_2$} contains unseen books. 
The participants can regard each chapter as a document to train and test their discourse-aware models. 
Apart from this, parallel training data in the General MT Task can also be used for data augmentation.
In the final testing stage, participants use their systems to translate the {\em official testing set} (Test$_{final}$). We select around 20 consecutive chapters from each book. Thus, we participants could treat all chapters within a book as a long document\footnote{The participants can still regard one chapter as a document, which depends on the models' length capability.}. As seen, the document length of Test$_{final}$ is quite longer than other sets. The final testset contains two references: Reference 1 is translated by human translators and Reference 2 is bult by manually aligning bilingual text in web page.
The genres in the valid and test sets are sampled evenly.

\subsection{Pretrained Models}
\label{sec:2-2}

Apart from training dataset from web novels, we also provide in-domain pretrained models as supplementary resources. These models can be used to finetune or initialize MT models. 
\begin{itemize}[leftmargin=*,topsep=0.1em,itemsep=0.1em,parsep=0.1em]

\item {\bf RoBERTa (base)}: The original model features a 12-layer encoder and is trained on the Chinese Wikipedia~\cite{liu2019roberta}. It has a hidden size of 768 and a vocabulary size of 21,128 using whole word masking. We continuously train it with Chinese literary texts (84B tokens)~\cite{wang2023disco}.

\item {\bf mBART (CC25)}: This original model is equipped with a 12-layer encoder and a 12-layer decoder, having been trained on a web corpus spanning 25 languages~\cite{liu2020multilingual}. It boasts a hidden size of 1024 and a vocabulary size of 250,000. We continuously train it with English and Chinese literary texts (114B tokens)~\cite{wang2023disco}. 

\end{itemize}
\noindent Besides, general-domain pretrained models listed in General MT Track are also allowed in this task: mBART, BERT, RoBERTa, sBERT, LaBSE.

\section{Evaluation Methods}
\label{sec:3}

It is still an open question whether human and automatic evaluation metrics are complementary or mutually exclusive in measuring the document-level and literary translation quality. Thus, we report both automatic and human evaluation methods, and officially rank the systems based on the overall human judgments.

\subsection{Automatic Evaluation}

We use widely-used sentence- and document-level evaluation metrics: 1) {\em sentence-level}: we employ sacreBLEU \cite{post2018call}, chrF \cite{popovic2015chrf},  TER~\cite{snover2005study} and pretraining-based COMET \cite{rei2020comet}; 2) {\em document-level}: we mainly use document-level sacreBLEU (d-BLEU) \cite{liu2020multilingual}, which is computed by matching n-grams in the whole document. For d-BLEU, We combine all sentences in each document as one line and then conduct sacreBLEU metric. Note that all evaluations are case-sensitive. 
We employ {\em sacrebleu}\footnote{\url{https://github.com/mjpost/sacrebleu} with signature: \texttt{nrefs:2|case:mixed|eff:no|tok:13a|smooth:exp |version:2.3.1}.} to calculate sacreBLEU, chrF, TER and d-BLEU with {\em sacrebleu} using two references.  The command is: \texttt{cat output | python -m sacrebleu reference*}.
We employ {\em unbabel-comet}\footnote{\url{https://github.com/mjpost/sacrebleu}.} to calculate COMET score using {\em Reference 1}. The command is: \texttt{comet-score -s input -t output -r reference1} (default model).

\subsection{Human Evaluation}

The human evaluation was performed by professional translators using an adaptation of the multidimensional quality metrics (MQM) framework \cite{lommel2014multidimensional}. For example, we consider the preservation of literary style and the overall coherence and cohesiveness of the translated texts. As shown in Table~\ref{tab:evaluation_criteria}, we put forth an industry-endorsed criteria to guide human evaluation process. The main error types are:
\begin{itemize}[leftmargin=*,topsep=0.1em,itemsep=0.1em,parsep=0.1em]

\item \textbf{Accuracy (Acc.)}: The target text does not accurately reflect the source text, allowing for any differences authorized by specifications.

\item \textbf{Fluency (Flu.)}: Issues related to the form or content of a text, irrespective as to whether it is a translation or not.

\item \textbf{Style (Sty.)}: The text has stylistic problems.

\item \textbf{Terminology (Ter.)}: A term (domain-specific word) is translated with a term other than the one expected for the domain or otherwise specified.

\item \textbf{Locale Convention (Loc.)}: The text does not adhere to locale-specific mechanical conventions and violates requirements for the presentation of content in the target locale.

\item \textbf{Others (Oth.)}: Other issues such as the signs of MT, gender bias and source errors.

\end{itemize}

MQM utilizes a scorecard format to quantify the quality assessment results. Evaluators assign numerical values to identified translation errors based on error types, severity, etc., making the assessment results more intuitive. The overall quality score is calculated based on per-word translation accuracy:

\begin{equation}
\nonumber
\text{S} = 1 - \frac{5 \times C_{\text{Min.}} + 10 \times C_{\text{Maj.}} + 25 \times C_{\text{Cri.}}}{\text{Total Word Count}}
\end{equation}
\noindent where where we set four error severity levels: Neutral (Neu.), Minor (Min.), Major (Maj.), Critical (Cri.) with 0/5/10/25 severity penalty. $C_{\star}$ denotes the number of errors. The ``Total Word Count'' is calculated based on source input (Chinese word).
Considering our task is centered on Zh-to-En translation, we engaged four evaluators who are native English speakers and also fluent in Chinese.

\section{Task Description}
\label{sec:4}
\paragraph{Overview}

The shared task will be the translation of literary texts between Chinese$\rightarrow$English. 
Participants will be provided with two types of training datasets: (1) discourse-level GuoFeng Webnovel Corpus; (2) General MT Track Parallel Training Data. Additionally, they are provided two types pretrained models: (1) in-domain pretrained models, including In-domain RoBERTa (base) and In-domain mBART (CC25). (2) other general-domain pretrained models listed in General MT Track. Note that basic linguistic tools are allowed in the constrained condition as well as pretrained language models released before February 2023.

In the final testing stage, participants use their systems to translate an official testing set. The translation quality is measured by a manual evaluation and automatic evaluation metrics. All systems will be ranked by human judgement according to our professional guidelines and translators. Participants can submit either constrained (i.e. only use the training data specified above) or unconstrained (i.e. it allows the participation with a system trained without any limitations) systems with flags, and we will distinguish their submissions.

\paragraph{Goals} 

The {main goals} of the task are to:
\begin{itemize}[leftmargin=*,topsep=0.1em,itemsep=0.1em,parsep=0.1em]
\item Encourage research in machine translation for literary texts.
\item Provide a platform for researchers to evaluate and compare the performance of different machine translation systems on a common dataset.
\item Advance the state of the art in machine translation for literary texts.
\end{itemize}

\paragraph{Submission and Format}

Submissions will be done by sending us an email to our official email. Each team can submit at most 3 MT outputs per language pair direction, one primary and up to two contrastive. The requirements of submission format are (1) Keep 12 output files that are identical to the testing input files. (2) In the output files, ensure that each line is aligned with the corresponding input line.

\begin{table*}[t]
    \centering
    \scalebox{0.92}{
    \begin{tabular}{c c c c c c}
     \toprule
     \bf ID & {\bf Team} & {\bf Institution} & \bf Flag & \bf \#System & \bf Main Methods\\
    \midrule
   1  & MaxLab & University of Southern California & $\bigodot$ & 3 & para-level Transformer\\
   2  & MAKE-NMT-VIZ & Université Grenoble Alpes & $\bigodot$ & 3 &  mBART\\
   3  & TJUNLP & Tianjin University & $\bigodot$ & 1 & sent-level Transformer\\
   4  & DLUT & Dalian University of Technology & $\bigotimes$ & 1 & GPT-3.5-turbo\\
   5  & NTU & Nantong University & $\bigotimes$ & 1 & Opus-MT\\
   6  & HITer-WMT & Harbin Institute of Technology & $\bigotimes$ & 2 & Llama-7b\\
   7  & HW-TSC & Huawei Translation Services Center & $\bigotimes$ & 3 & doc2doc Transformer\\
    \bottomrule
    \end{tabular}}
    \caption{The summary of system submission and their participant teams. We also report the number of systems (\#System) and the constrained ($\bigodot$) and unconstrained ($\bigotimes$) flags. }
    \label{tab:sys-summary}
\end{table*}

\section{Participants' and Baseline Systems}
\label{sec:5}

Here we briefly introduce each participant's systems and refer the reader to the participant's reports for further details. Table \ref{tab:sys-summary} shows the summary of systems and participant teams.

\subsection{MaxLab (constrained)}

The team from University of Southern California, Information Sciences Institute introduce three translation systems. 
The {\em Primary System} is built on a paragraph-level transformer, trained on a paragraph-aligned corpus (with a source side cap of 256 characters), executing translations at the paragraph level.
The {\em Contrastive System 1} deploys a sentence-level transformer, capitalizing on the sentence alignment data available in the datasets. 
The {\em Contrastive System 2} adopts a paragraph-level Mega model \cite{ma2022mega}. The Mega model proposed a single-head gated attention mechanism equipped with an exponential moving average, which achieves comparable performance compared to Transformers having with fewer parameters. 
In pre-processing, the team opted for Byte-Pair Encoding (BPE) for tokenization. And they employed Jaccard similarity for sentence alignment during the post-processing phase.

\subsection{MAKE-NMT-VIZ (constrained)}

The team from Université Grenoble Alpes introduced three translation systems. 
The {\em Primary System} finetune the mBART (CC50) model using Train, Valid$_1$, Test$_1$ of the GuoFeng Corpus, adopting settings similar to those described by \citet{lee2022pre}. Specifically, they finetune models for 3 epochs, utilizing the GELU activation function, a learning rate of 0.05, a dropout rate of 0.1, and a batch size of 16. For decoding, a beam search of size 5 was employed.
The {\em Contrastive System 1} is implemented upon a finetuned concatenation transformer \cite{lupo2023encoding} with two training steps: (1) a sentence-level transformer is trained for 10 epochs using General, Valid$_1$, Test$_1$ datasets; (2) a document-level transformer is finetuned using pseudo-document data (3-sentence concatenation) from Train, Valid$_2$, Test$_2$ data for 4 epochs. They use ReLU as an activation function, along with an inverse square root learning rate, a dropout rate of 0.1, and a batch size of 64. For decoding, a beam search of size 4 was employed.
The {\em Contrastive System 2} is a sentence-level transformer model trained for 10 epochs using General, Valid$_1$, Test$_1$ datasets. The training adopted an inverse square root scheduled learning rate, a dropout rate of 0.1, and a batch size of 64. Decoding was done using a beam search of size 4.

\subsection{TJUNLP (constrained)}

The team from Tianjin University introduced a {\em Primary System} based on a sentence-level Transformer model. The training consists of two phases: initially, it undergoes 100k steps on a dense model, followed by a 50k step fine-tuning on mixture of experts (MOE). They adopt the Polynomial Decay as their learning rate scheduling strategy, with a learning rate set at 2e-4, a dropout rate of 0.1, and a batch size encompassing 4096 tokens. For decoding, a beam search of size 5 was employed. For pre-processing, the team opted for SentencePiece Model (SPM) for tokenization.

\subsection{NTU (unconstrained)}

The Nantong University team introduce a {\em Primary System}. It is based on a pretrained MT model, Opus-MT,\footnote{\url{https://huggingface.co/Helsinki-NLP/opus-mt-zh-en}.}, which is trained on OPUS dataset \cite{opus}. The model is finetuned on one NVIDIA Tesla A100 80 GB where the learning rate is {5e-5}, batch size is 64, max length is 512 and the epoch number is 10.

\subsection{DLUT (unconstrained)}

The team form Dalian University of Technology introduce a {\em Primary System} based on GPT-3.5-turbo \cite{brown2020language}. They mainly propose prompt engineering, data filtering, and document segmentation to activate the capabilities of LLMs for discourse-level translation \cite{zhao-etal-2023-DLUT-system}.

\subsection{HITer-WMT (unconstrained)}

The team form Harbin Institute of Technology (Harbin) introduce two translation systems. 
The {\em Primary System} centers on instruction fine-tuning, executed through the Llama-7b model within the Parrot framework~\cite{jiao2023parrot}.\footnote{\url{https://github.com/wxjiao/ParroT}.} Specifically, they build an instruction dataset from two comprehensive chapters of our existing training corpus according to methodologies in \citet{peng2023instruction}. This dataset was fine-tuned using Llama-7b over 3 epochs with a learning rate of 2e-5.
The {\em Contrastive System} utilizes the GuoFeng mBART Model provided by the shared task. This model was trained over 10 epochs at a learning rate of 1e-4, with gradient clipping applied to stabilize training. 

\begin{table*}[t]
    \centering
    \scalebox{0.94}{
    \begin{tabular}{cc rrrrr}
     \toprule
     \multirow{2}{*}{\bf Type} & \multirow{2}{*}{\bf System} &  \multicolumn{4}{c}{\color{blue} Sent-Level}  & \multicolumn{1}{c}{\color{red} Doc-Level} \\
     \cmidrule(lr){3-6} \cmidrule(lr){7-7} 
     & & \bf BLEU$^\uparrow$ & \bf chrF$^\uparrow$ & \bf COMET$^\uparrow$ & \bf TER$^\downarrow$ & \bf d-BLEU$^\uparrow$ \\
    \midrule
   \multirow{3}{*}{\em Baselines} & Llama-MT$^\star$ & n/a & n/a & n/a &  n/a& 43.1\\
    & GPT-4$^\star$ & n/a & n/a & n/a & n/a & 43.7\\
    & Google$^\star$ & 37.4 & 57.0 & 80.50 & 57.4 & 47.3\\
    \midrule
    \multirow{3}{*}{\em \shortstack{Primary \\ (con)}} & MaxLab & 34.1 & 53.3 & 78.24 & 62.4& 45.0\\
    & MAKE-NMT-VIZ & {\color{blue}37.9} & {\color{blue}56.6} & {\color{blue}81.50} & {\color{blue}58.7} & \bf {\color{red}48.0}\\
    & TJUNLP & 32.1 & 51.9 & 77.93 & 64.1 & 43.3\\
    \cmidrule{1-7}
    \multirow{4}{*}{\em \shortstack{Primary \\ (uncon)}} & DLUT$^\star$ & 40.5 & 58.5 & 82.58 & 54.6 & 50.2 \\
    & NTU$^\star$  & 32.3 & 52.5 & 78.07 & 64.3 & 43.4 \\
    & HITer-WMT$^\star$ & 16.1 & 37.1 & 69.84 & 80.1 & 28.0 \\
    & HW-TSC$^\star$ & {\color{blue}44.3} & {\color{blue}61.1} & {\color{blue}82.69} & {\color{blue}51.8} & \bf {\color{red}52.2}\\
    \midrule
    \multirow{7}{*}{\em Contrastive} & MaxLab$_{1}$ & 34.5 & 54.7 & 79.14 & 62.7 & 44.9\\
    & MaxLab$_{2}$ & 33.1 & 52.4 & 77.84 & 63.6 & 44.4\\
    \cdashline{2-7}\noalign{\vskip 0.5ex}
    & MAKE-NMT-VIZ$_{1}$  & 33.8 & 51.2 & 76.91 & 63.5 & 45.5\\
    & MAKE-NMT-VIZ$_{2}$  & 35.0 & 52.7 & 77.26 & 61.5 & 46.2\\
    \cmidrule{2-7}
    & HITer-WMT$_{1}^\star$  & 30.8 & 49.2 & 76.41 & 67.2 & 40.6\\
    \cdashline{2-7}\noalign{\vskip 0.5ex}
    & HW-TSC$_{1}^\star$ & 44.6 & 61.0 & 82.67 & 51.8 & 52.6\\
    & HW-TSC$_{2}^\star$ & 44.4 & 61.5 & 82.63 & 52.1 & 52.2\\
    \bottomrule
    \end{tabular}}
    \caption{Evaluation results of baseline and participants' systems in terms of {\bf automatic evaluation methods}, including 1) {\color{blue}sentence-level} metrics BLEU, chrF, COMET, TER; and 2) {\color{red}document-level} metrics d-BLEU. Systems marked with $^\star$ are unconstrained, while others are constrained. The COMET is calculated with {\em unbabel-comet} using {\em Reference 1} while others are calculated with {\em sacrebleu} using two references. The best primary constrained and unconstrained systems are highlighted.}
    \label{tab:auto-res}
\end{table*}

\subsection{HW-TSC (unconstrained)}

The team form Huawei Translation Services Center exploit a variety of techniques. 
They introduce an unconstrained Document-to-Document Translation system. They first train a sentence-level Transformer-big model with a 25-layer encoder and a 6-layer decoder, and perform domain adaptation with novel data on this model. They obtain a strong baseline using data augmentation methods including Back Translation, Forward Translation, and Data Diversification. They then perform incremental training using the Doc2Doc technique to turn the model into a document-level translation model. They also conduct document-level data augmentation using the Multi-resolutional Document-to-Document approach \cite{sun2022rethinking}, and ensue the consistency of NE translations in a document with TrAining Data Augmentation (TADA). They submit three systems: the {\em Primary System} uses all strategies. In contrast to the primary system, the {\em Contrastive System 1} system does not use TADA, and the {\em Contrastive System 2} sets the beam size to 6 during inference, while 10 for other tasks.

\subsection{Baseline Systems (unconstrained)}

We select three representative systems as baselines. \textit{Commercial Translation System}: we use Google Translate,\footnote{\url{https://translate.google.com}.}, which usually performs state-of-the-art in translation performance. \textit{Commercial LLM Systems}: we employ GPT-4 (8K) API\footnote{\url{https://platform.openai.com}.} to translate documents, which is known for its extensive context modeling capabilities \cite{ouyangtraining-instructGPT,wang2023document}. \textit{Open-sourced LLM Models}: we enhance Llama (2K)~\cite{touvron2023llama} on document-level translation by using the 200K general-domain document-level training set \cite{du2023on}. 
All testing were conducted between August 1st and 30th, 2023.
In the future, we will use more diverse model architectures such as non-autoregressive translation model \cite{gu2017non,ding2020understanding,ding2021progressive,wang2023revisiting}.

\begin{table}[t]
    \centering
    \scalebox{0.95}{
    \begin{tabular}{c c r c}
     \toprule
     {\bf Type} & {\bf System} & {\color{red} MQM} & {\color{forestgreen}Rank}\\
    \midrule
   \multirow{3}{*}{\em Baselines} & GPT-4$^\star$ & 54.81 & 1\\
   & Llama-MT$^\star$ & 28.40  & 2\\
   & Google$^\star$ & 22.66  & 3\\
    \cmidrule(lr){1-4}
    \multirow{3}{*}{\em \shortstack{Primary \\ (con)}} & MAKE-NMT-VIZ & {\color{red} \bf 42.36} & {\color{forestgreen} \bf 1}\\
    & MaxLab & 28.58  & 2\\
    & TJUNLP & 18.34  & 3\\
    \cmidrule{1-4}
    \multirow{4}{*}{\em \shortstack{Primary \\ (uncon)}} & DLUT$^\star$ & {\color{red} \bf 63.35} & {\color{forestgreen} \bf 1}\\
    & HW-TSC$^\star$ & 53.01 & 2\\
    & NTU$^\star$  & 31.66 & 3\\
    & HITer-WMT$^\star$ & 5.56 & 4\\ 
    \bottomrule
    \end{tabular}}
    \caption{Evaluation results of baseline and primary systems in terms of {\bf human evaluation}. We report {\color{red}MQM score} and {\color{forestgreen} System Rank}.}
    \label{tab:human-res}
\end{table}

\begin{table*}[t]
    \centering
    \scalebox{0.94}{
    \begin{tabular}{cc rrrr r}
     \toprule
     \multirow{2}{*}{\bf Type} & \multirow{2}{*}{\bf Systems} &  \multicolumn{4}{c}{\color{blue} Annotator}  & \multirow{2}{*}{\color{red} Average} \\
     \cmidrule(lr){3-6}
     & & \bf 1 & \bf 2 & \bf 3 & \bf 4 \\
    \midrule
   \multirow{3}{*}{\em Baselines} & GPT-4$^\star$ & 95.84 & 73.38 & 76.71 & 87.52 & 83.36 \\
   & Llama-MT$^\star$ & 94.18 & 65.06 & 78.37 & 83.36 & 80.24 \\
    & Google$^\star$ & 85.02 & 42.60 & 59.23 & 21.13 & 52.00 \\
    \cmidrule(lr){1-7}
    \multirow{3}{*}{\em \shortstack{Primary \\ (con)}} & MAKE-NMT-VIZ & {\color{blue}97.50} & {\color{blue}83.36} & {\color{blue}92.51} & {\color{blue}91.68} & {\color{red}\bf 91.26} \\
    & MaxLab & 86.69 & 61.73 & 71.71 & 74.21 & 73.59 \\
    & TJUNLP & 88.02 & 55.07 & 20.97 & 69.22 & 58.32 \\
    \cmidrule{1-7}
    \multirow{4}{*}{\em \shortstack{Primary \\ (uncon)}} & HW-TSC$^\star$ & 91.68 & {\color{blue}83.36} & {\color{blue}83.36} & {\color{blue}91.68} & {\color{red}\bf 87.52} \\
    & DLUT$^\star$ & {\color{blue}95.01} & 69.22 & 84.19 & 90.02 & 84.61 \\
    & NTU$^\star$  & 85.02 & 39.27 & 28.45 & 62.56 & 53.83 \\
    & HITer-WMT$^\star$ & 57.57 & 21.80 & 0.00 & 31.78 & 27.79 \\
    \bottomrule
    \end{tabular}}
    \caption{Analysis of human scores by different annotators on {\bf one sampled document}. We report {\color{blue}four annotators' scores} and {\color{red}average score} of Baselines, primary constrained and unconstrained ($^\star$) systems. }
    \label{tab:human-abl}
\end{table*}

\begin{table}[t]
\centering
\scalebox{0.95}{
  \begin{tabular}{c c c c c c}
     \toprule
     {\color{blue} Annotator} & \bf 1 & \bf 2 & \bf 3 & \bf 4\\
     \midrule
     \bf 1 & - & - & - & - \\
     \bf 2 & 0.858 & - & - & - \\ 
     \bf 3 & 0.824 & 0.878 & - & - \\
     \bf 4 & 0.752 & 0.875 & 0.676 & - \\
     {\color{red} Average} & 0.902 & 0.976 & 0.927 & 0.891 \\
    \bottomrule
    \end{tabular}}
    \caption{Pearson correlation coefficient between scores by different annotators in Table \ref{tab:human-abl}.}
    \label{tab:human-analysis}
\end{table}


\section{Evaluation Results}
\label{sec:6}

\subsection{Automatic Evaluation}

We report the automatic evaluation scores of all submissions in Table \ref{tab:auto-res}. The evaluation metrics includes 1) sentence-level BLEU, chrF, COMET, TER; and 2) document-level d-BLEU. To calculate d-BLEU, we first concatenate all continuous sentences in one book as on line, and then employ sacreBLEU to obtain scorers. To compute d-BLEU, we merge all the consecutive sentences from a single book into one continuous line, and then utilize the sacreBLEU to generate the scores.

Among constrained Primary systems, the MAKE-NMT-VIZ system shows impressive performance and achieves the best in terms of all metrics. Similarly, the HW-TSC$^\star$ Primary system achieves the best in constrained settings. As introduced in Section \ref{sec:5}, MAKE-NMT-VIZ mainly finetune the mBART pretrained model while HW-TSC$^\star$ train a doc2doc Transformer model using a number of data augmentation methods.

In the majority of teams, the primary system exhibits superior performance compared to the corresponding contrastive system. The exceptions to this trend are noted in the cases of HITer-WMT$^\star$ and HW-TSC$^\star$, where this pattern does not hold.
Among the baseline systems, Google Translate, a commercial translation service, outperforms both commercial and open-source LLMs (GPT-4 API and Llama-MT) in terms of d-BLEU scores. Interestingly, both the top-1 ranked Primary constrained and the top-2 ranked unconstrained systems surpass the performance of the commercial MT system.

\subsection{Human Evaluation}

Table \ref{tab:human-res} presents the results of the human evaluation and system rank for the Primary submissions. We enlisted four human annotators to evaluate 5 documents, comprising a total of 2,194 words sourced from distinct books within the final testset for each translation system.

As seen, the MAKE-NMT-VIZ system outperforms the other three constrained systems, while DLUT$^\star$ ranks first among the four unconstrained systems.
This is not fully consistent with the automatic evaluation results in Table \ref{tab:auto-res}. Moreover, the top-2 unconstrained systems outperform the best constrained system, highlighting the benefits of external knowledge. This observation is consistent with that of automatic evaluation. 

Among the baseline systems, the LLM system performs the best, whereas the MT system shows the poorest performance, diverging from the observations of automatic evaluation. Interestingly, the literary MT-enhanced models perform comparable with some systems such as MaxLab and Google Translate.

\subsection{Analysis}

\paragraph{Inter-Annotator Agreement}

We engaged four annotators to independently review an identical document (i.e. 601 words) selected from the testset. Table \ref{tab:human-abl} outlines the individual scores given by each annotator and the corresponding average scores. The findings illustrate that (1) while there is variance in the exact scores assigned by different annotators, their scoring trends align; (2) the results on this sample may diverge from those obtained from a larger dataset, highlighting the necessity of human evaluation on a larger scale.

In our effort to understand the consistency among the human evaluators, we conducted a Pearson correlation analysis on their scoring patterns. Table~\ref{tab:human-analysis} illustrates the pairwise Pearson correlation coefficients for the scores given by each annotator.
The results indicate a high degree of agreement among the annotators. For example, Annotator 2 demonstrated a very high correlation with Annotator 3 (\(r = 0.878\)) and Annotator 4 (\(r = 0.875\)). Besides, the Average Scores also reveal strong evaluator consensus on translation quality. This consistency underscores the reliability of the evaluators' judgments across the assessed translations.

\begin{figure*}[t] 
    \centering
    \includegraphics[width=1\textwidth]{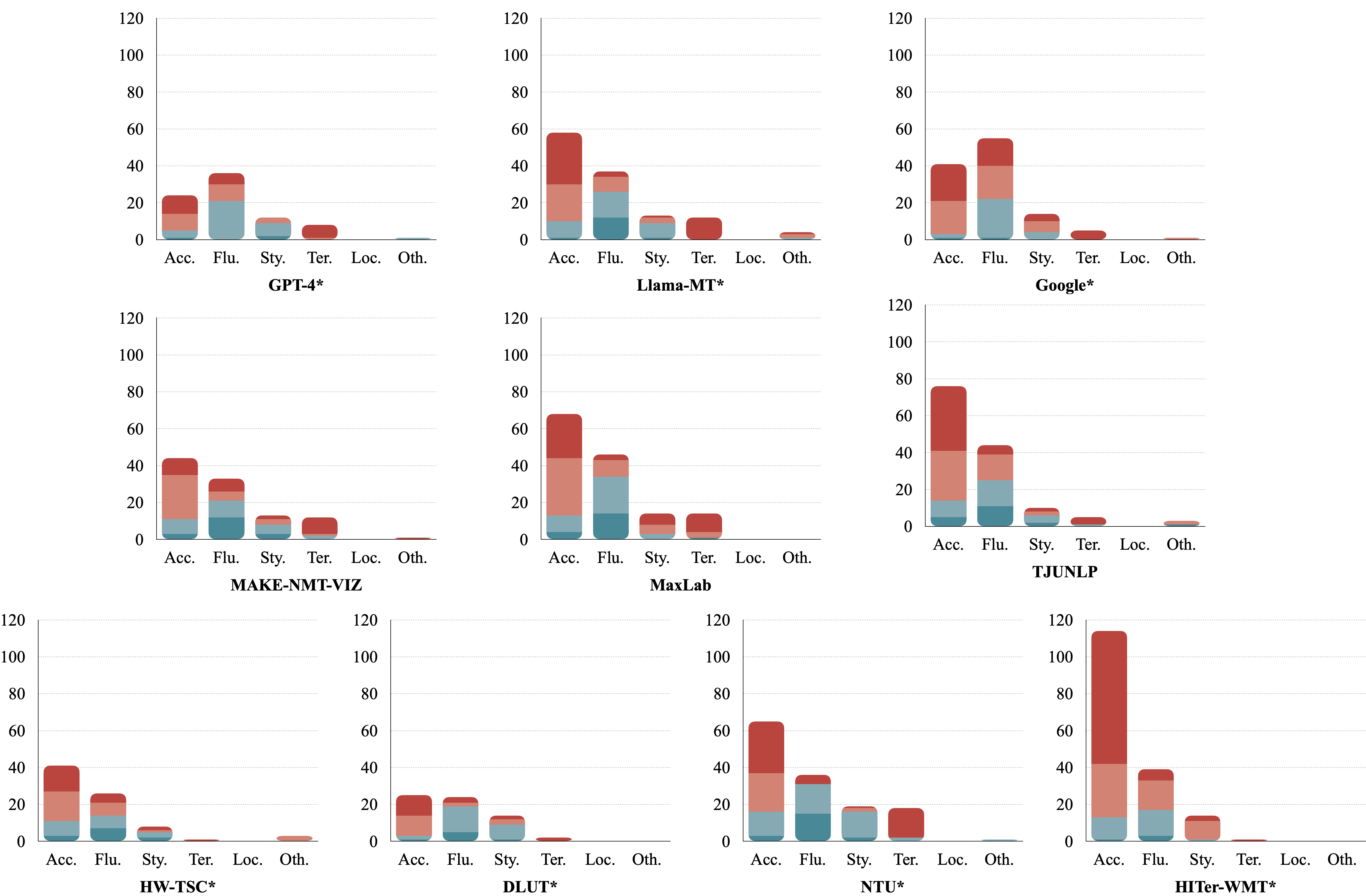}
    \caption{Analysis of error types in human annotations: Accuracy (Acc.), Fluency (Flu.), Style (Sty.), Terminology (Ter.), Localization (Loc.), and Other (Oth.). We report the count of error checkpoints in four evaluated documents. The four error severity levels are presented in different colors: Neutral (blue), Minor (light blue) , Major (light red), Critical (red). Systems marked with $^\star$ are unconstrained, while others are constrained.} 
    \label{fig:error-type}
\end{figure*}

\paragraph{Error Type}

We further analyze the error distribution in human-annotated results. Figure \ref{fig:error-type} classifies and counts the errors identified in the evaluated documents by their severity. This visualization allows for a direct comparison of the error profiles of each system, highlighting their strengths and weaknesses in different aspects of translation quality.

In the baseline systems analysis, GPT-4$^\star$ registers a higher frequency of Minor errors, particularly in Fluency and Style, indicating areas where refinement could enhance the translation's naturalness and adherence to stylistic norms. Llama-MT$^\star$, by contrast, has a pronounced incidence of Major and Critical errors in Accuracy and Terminology, raising concerns about the fidelity and technical precision of its translations. Google$^\star$ stands out with its Fluency errors, suggesting potential issues in maintaining a coherent and natural flow compared to the language models.

Regarding the constrained systems, MAKE-NMT-VIZ displays an even spread of errors, with relatively fewer instances in each category, which points to a well-rounded performance in capturing nuances across various aspects of translation. Both MaxLab and TJUNLP exhibit an increased number of Accuracy and Fluency errors, suggesting challenges in delivering translations that are not only faithful to the source material but also exhibit a seamless and natural flow in the target language.

The unconstrained systems, particularly HW-TSC$^\star$ and DLUT$^\star$, show a notable reduction in errors related to Accuracy and Fluency when compared to their constrained counterparts. This trend suggests that the lack of constraints may afford these systems more flexibility, resulting in translations that are more accurate and fluid. However, the overall error distribution across different systems highlights the complex trade-offs and challenges inherent in machine translation, underscoring the need for continued innovation and optimization in the field. In the future, we will also consider hallucination errors \cite{zhang2023siren}.

\section{Conclusion and Future Work}
\label{sec:7}

We believe that the WMT2023 Shared Task on discourse-level literary translation will be a valuable contribution to the field of machine translation and will encourage further research in this area. We discuss the potential limitations of this edition of the shared task as follows:
\begin{itemize}[leftmargin=*,topsep=0.1em,itemsep=0.1em,parsep=0.1em]

\item {\em Language Pair}. This year, we only focus on Chinese$\rightarrow$English direction. However, we have a long-term plan to continuously organize this task, and will extend the copyrighted dataset into Chinese-Russian and Chinese-German language pairs next year. 

\item {\em Literary Genre}. This year, we mainly used the Web Fiction Corpus which is only one type of literary text. We use Web Fiction for two reasons: (1) its literariness is less complicated than others (e.g. poetry, masterpiece); (2) such bilingual data are numerous and continuously increased. We will consider to extend more literary genres such as poetric translation in the next year.

\item {\em Discourse Benchmark}. We have accumulated some discourse- and context-aware benchmarks \cite{xu-etal-2022-guofeng,xu2023benchmark,wang2023disco}. These benchmarks are pivotal for assessing the proficiency of LLMs in handling complex language structures and contextual nuances. As participation of LLM-based systems in our shared tasks increases, we anticipate integrating these benchmarks more comprehensively into our future evaluations to better measure and understand the evolution of LLM capabilities in linguistic context and discourse comprehension.

\end{itemize}

Machine translation of web novels not only holds research value but also offers practical application prospects \cite{huang2021transmart,lyu2023new}. 
This shared task serves to spur competitive innovation and fosters the advancement of sophisticated machine translation systems capable of navigating the intricate nuances of literary works.
Anticipating the future, our objective is to broaden the engagement in the forthcoming shared task, inviting an extensive range of collaborators from industry and academia alike to contribute their unique insights and expertise.


\section*{Acknowledgements}
We would like to thank the WMT2023 organizers for providing us the opportunity to explore this new task.
We also express our gratitude to the experts on the Shared Task Committee for their efforts in organization, evaluation, and advisory roles:
\begin{itemize}[leftmargin=*,topsep=0.1em,itemsep=0.1em,parsep=0.1em]
\item Longyue Wang, Zhaopeng Tu, Dian Yu, Chenyang Lyu, Shuming Shi (Tencent AI Lab)
\item Yan Gu, Yufeng Ma, Weiyu Chen (China Literature Ltd.)
\item Bonnie Webber (University of Edinburgh)
\item Siyou Liu, Yulin Yuan (University of Macau)
\item Philipp Koehn (Johns Hopkins University)
\item Liting Zhou, Andy Way (Dublin City University)
\item Yvette Graham (Trinity College Dublin)
\item Chao-Hong Liu (Potamu Research Ltd.)
\item Qingsong Ma (Tencent AI Evaluation Lab)
\end{itemize}

\begin{CJK*}{UTF8}{gbsn}

\begin{table*}[h]
    \centering
    \small
    \scalebox{1}{
    \begin{tabular}{p{1.6cm} p{1.9cm} p{4.3cm} p{6.5cm}}

        \toprule
        \textbf{Type} & \textbf{Granular} & \textbf{Definition} & \textbf{Examples} \\
        \midrule
        
        \multirow{20}{*}{\bf Accuracy} & Addition & The target text includes text not present in the raw. & A translation includes portions of another translation that were inadvertently pasted into the document or the translator has added too many details of his own. \\\cdashline{2-4}\noalign{\vskip 0.5ex}
        
        & Omission & Content is missing from the translation that is present in the source. & A paragraph present in the source is missing in the translation. \\\cdashline{2-4}\noalign{\vskip 0.5ex}
        
        & Mistranslation & The target content does not accurately match the raw. & A source text states that a medicine {\em should not be} administered in doses greater than 200 mg, but the translation states that it should be administered in doses greater than 200 mg (i.e., negation has been omitted). \\\cdashline{2-4}\noalign{\vskip 0.5ex}
        
        & Misnomer & The target text is more/less specific than the raw. & 1. The source text refers to a boy but is translated with a word that applies only to young boys rather than the more general term. 2. The source text uses words that refer to a specific type of military officer but the target text refers to military officers in general. \\\cdashline{2-4}\noalign{\vskip 0.5ex}
        
        & Untranslated & Content that should have been translated has been left untranslated. & A sentence to be translated into English was left in Chinese. \\
        
        \midrule
        \multirow{14}{*}{\bf Fluency} & Punctuation & Punctuation marks missing or used in a wrong way. & An English text uses a semicolon where a comma should be used. \\\cdashline{2-4}\noalign{\vskip 0.5ex}
        
        & Spelling & Issues related to spelling of words. (Including those of capitalization, hyphenated words, and use of asterisk for censored swear words.) & The English word ``Translation'' is spelled ``Transaltion''. \\\cdashline{2-4}\noalign{\vskip 0.5ex}
        
        & Grammar & Issues related to the grammar or syntax of the text, other than spelling and orthography. (especially inconsistency of the tenses and conditionals) & An English text reads ``The man was seeing the his wife.'' \\\cdashline{2-4}\noalign{\vskip 0.5ex}
        
        & Inconsistency & The text shows internal inconsistency. & A text uses both ``app.'' and ``approx.'' for ``approximately''. \\
        
        \midrule
        \multirow{12}{*}{\bf Style} & Awkwardness & A text is written with an awkward style. & A text is written with many embedded clauses and an excessively wordy style. While the meaning can be understood, the text is very awkward and difficult to follow. \\\cdashline{2-4}\noalign{\vskip 0.5ex}
        
        & Inconsistent & Style is inconsistent within a text. & One part of a text is written in a light and terse style while other sections are written in a more wordy style. \\\cdashline{2-4}\noalign{\vskip 0.5ex}
        
        & Unidiomatic & The content is grammatical, but not idiomatic. & The following text appears in an English translation of ``我们衷心感谢他'': ``We thanked him with heart'' where ``with heart'' is an understandable, but non-idiomatic rendering, better stated as ``heartily''.  \\
        
        \midrule
        \multirow{6}{*}{\bf Terminology} & Mistranslation & A genre-specific or cultural-specific terminology is wrongly translated. & A Chinese word ``修士'' is translated into ``practitioner'' rather than the expected ``cultivator''. \\\cdashline{2-4}\noalign{\vskip 0.5ex}
        
        & Inconsistent & Terminology is used in an inconsistent manner within the text. & ``斗罗大陆'' is translated into ``Douluo Land'' in the first few chapters and then into ``Soul Land''. \\
        
        \midrule
        \multirow{8}{*}{\makecell{\bf Locale \\ \bf Convention}} & Location Format & Using the wrong format for address, name etc. & A Chinese address ``北京市朝阳区花园路22号'' is translated into ``Beijing, Chaoyang district, Huayuan Road N.22'' instead of the expected ``N.22, Huayuan Road, Chaoyang District, Beijing''. \\\cdashline{2-4}\noalign{\vskip 0.5ex}
        
        & Number Format & The translated date, time, currency, telephone use formats inappropriate for its locale. & An English text has 2012-06-07 instead of the expected 06/07/2012. \\
        
        \midrule
        
        \multirow{2}{*}{\bf Others} &  & Other issues that haven't been included in this list. &  E.g. signs of MT, mimetic word, gender bias, source errors etc.\\
        \bottomrule
    \end{tabular}
    }
    \caption{The MQM-based evaluation criteria for literary translation.}
    \label{tab:evaluation_criteria}
\end{table*}

\end{CJK*}

\bibliography{custom}
\bibliographystyle{acl_natbib}




\end{document}